\documentclass{article}
\usepackage[a4paper, total={7in, 9in}]{geometry}
\usepackage{graphicx} 
\usepackage[sorting=none]{biblatex}
\usepackage{booktabs}
\usepackage{bm}
\usepackage{amsmath}
\usepackage{amssymb}
\usepackage{physics}
\usepackage{multirow}
\usepackage{mathtools}
\usepackage{algpseudocode}
\usepackage{algorithm}
\usepackage[dvipsnames]{xcolor}
\usepackage{caption}
\usepackage{authblk}
\usepackage{forest}
\usepackage{subfiles}
\usepackage[pdftex]{pict2e}
\usepackage[toc,page]{appendix}
\DeclareMathOperator*{\argmin}{arg\,min}

\title{Symbolic Discovery of Stochastic Differential Equations\\with Genetic Programming}
\author[1*]{Sigur de Vries}
\author[1]{Sander W. Keemink}
\author[1]{Marcel A. J. van Gerven}
\date{}
\affil[1]{Department of Machine Learning and Neural Computing, Donders Institute for Brain, Cognition and Behaviour, Radboud University,  Nijmegen, the Netherlands}
\affil[*] {sigur.devries@donders.ru.nl}

\begin{document}

\maketitle

\begin{abstract}
    \noindent Automated scientific discovery aims to improve scientific understanding through machine learning. A central approach in this field is symbolic regression, which uses genetic programming or sparse regression to learn interpretable mathematical expressions to explain observed data. Conventionally, the focus of symbolic regression is on identifying ordinary differential equations. The general view is that noise only complicates the recovery of deterministic dynamics. However, explicitly learning a symbolic function of the noise component in stochastic differential equations enhances modelling capacity, increases knowledge gain and enables generative sampling. We introduce a method for symbolic discovery of stochastic differential equations based on genetic programming, jointly optimizing drift and diffusion functions via the maximum likelihood estimate. Our results demonstrate accurate recovery of governing equations, efficient scaling to higher-dimensional systems, robustness to sparsely sampled problems and generalization to stochastic partial differential equations. This work extends symbolic regression toward interpretable discovery of stochastic dynamical systems, contributing to the automation of science in a noisy and dynamic world.
\end{abstract}

\section{Introduction}
The goal of automated scientific discovery (ASD) is to develop and deploy machine learning for scientific research~\cite{rackauckas2020universal, wang2023scientific}. In other words, ASD focuses on gathering knowledge about unknown systems or processes from observed data through machine learning. Generally, ASD includes generating and validating hypotheses in a data-driven fashion, substantially speeding up the process compared to human efforts. Successes of ASD have been seen in material sciences~\cite{morgan2020opportunities}, drug discovery~\cite{schneider2018automating}, chemistry~\cite{coley2020autonomous} and many other fields. An important direction within ASD is system identification, which involves building mathematical models to learn underlying mechanics from data. Often these mathematical models are purely deterministic, which leads to limitations in settings with noise, either as part of the dynamics (aleatoric uncertainty) or through incomplete observations (epistemic uncertainty). Capturing noise in the mathematical models is attracting increasing interest, as it supports generative modelling and uncertainty quantification.

One approach to system identification is symbolic regression, which discovers symbolic structures to explain the relation between variables~\cite{makke2024interpretable, udrescu2020ai}. Compared to traditional black-box optimization methods like neural networks, symbolic regression results in interpretable models that not only capture the data but can also provide information about the underlying system. Symbolic regression is primarily performed with genetic programming (GP)~\cite{koza1994genetic, bongard2007automated, schmidt2009distilling} or sparse regression~\cite{brunton2016discovering}, but less traditional algorithms also start to show competitive performance~\cite{d2023odeformer, kim2020integration, biggio2021neural, jin2019bayesian}. GP optimizes the structure of computer programs represented as parse trees~\cite{koza1994genetic}, taking inspiration from evolutionary principles. This allows learning the parameters and structures of mathematical equations directly, reducing human bias and introducing interpretability.

\begin{figure}[!t]
    \centering
    \includegraphics[width=\linewidth]{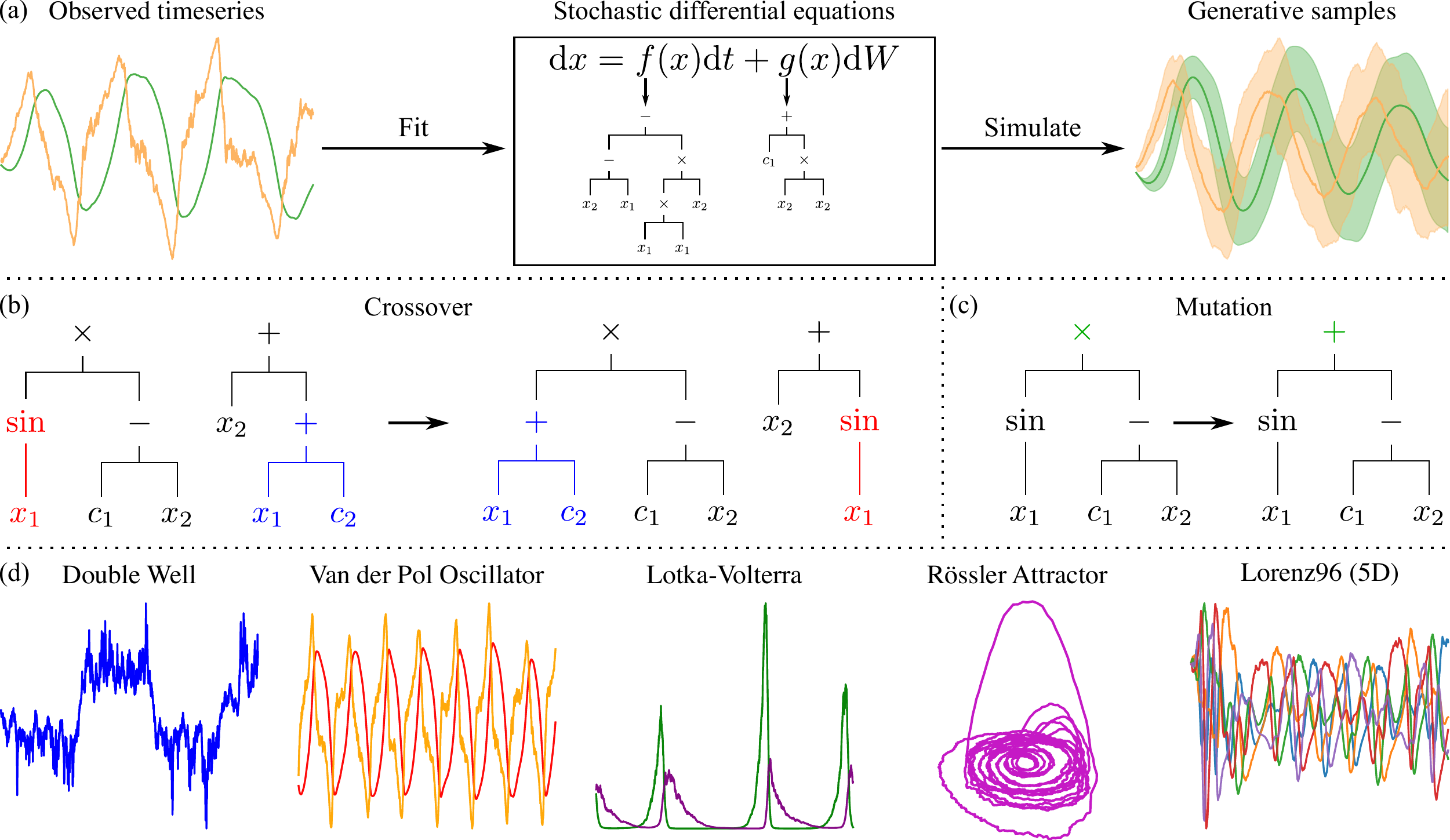}
    \caption{\textbf{Overview of genetic programming of stochastic differential equations.} (a) Stochastic time series are observed to which stochastic differential equations are fitted with symbolic expressions. The resulting stochastic differential equation offers interpretability and generative modelling. The symbolic equations for the drift $f(x)$ and diffusion $g(x)$ are optimized with genetic programming. (b) In genetic programming, trees are adapted with crossover that swaps subtrees, illustrated with the exchange of the red and blue subtrees. (c) Besides crossover, mutation also changes the structure of individual trees. Various aspects can be mutated, where this example shows the change of an operator in green. (d) Examples of systems on which the proposed method can be applied to. The curves in each subpanel illustrate the behaviour of the state following stochastic system dynamics. }
    \label{fig:overview}
\end{figure}

Most applications of symbolic regression focus on discovering governing laws or dynamical systems from data~\cite{schmidt2009distilling, udrescu2020ai,bongard2007automated, brunton2016discovering, quade2016prediction}. Within the field of dynamical systems, predominantly ordinary differential equations (ODEs) are learned, although partial differential equations (PDEs) have received growing interest in recent years~\cite{messenger2021weak, zhang2023deep, reinbold2019data}. Another class of dynamical systems are stochastic differential equations (SDEs)~(Fig.~\ref{fig:overview}a). An SDE consists of a deterministic function and additional noise processes that randomly perturb the system dynamics. Using SDEs in real world applications allows for generative modelling and stronger approximation~\cite{elgazzar2024generative, tzen2019neural, song2020score}. Learning the symbolic structure of SDEs remains relatively unexplored compared to ODEs and PDEs, and has yet to be investigated with GP. Noise in the system is typically considered an added challenge to quantify robustness of a method during identification of deterministic governing equations~\cite{brunton2016discovering, d2023odeformer}, but rarely the term that governs the noise is learned.

Currently, the only applications of symbolic regression to SDEs are performed with sparse regression, showcasing the identification of the correct structure of systems with Gaussian noise~\cite{boninsegna2018sparse, wang2022data, rajabzadeh2016robust}, L\'evy jumps~\cite{li2021data, li2025evolutionary}, and stochastic partial differential equations (SPDEs)~\cite{mathpati2024discovering}. By capturing the noise term, both the accuracy of learning the deterministic part of the dynamics improves and an interpretable function is discovered for the stochastic component of the dynamics, which provides an advantage over black-box approaches for fitting SDEs~\cite{rackauckas2020universal, tzen2019neural, jia2019neural}. To apply sparse regression to SDEs, the Kramers-Moyal expansion~\cite{risken1989fokker} generates separate estimates of the coefficients of the deterministic and the stochastic (diffusion) components of the dynamics using a binning approach. This imposes several limitations, such as that the drift and diffusion are not optimized simultaneously, inaccuracies can arise in both the expansion and regression and it is susceptible to the curse of dimensionality~\cite{honisch2011estimation}. 

Alternatively, GP could be extended to learn the structure of SDEs. Previous work has demonstrated that parameters could be fitted in (non-linear) SDEs with fixed structures by directly optimizing the maximum likelihood estimate (MLE)~\cite{nielsen2000parameter, ait2002maximum, pedersen1995new, kristensen2004parameter}. Hence, the MLE would also be a suitable objective function in GP to learn symbolic expressions for SDEs, jointly evolving the drift and diffusion using a multi-tree representation, while also removing the reliance on data binning and coefficient estimation. 

In this paper, we aim to apply GP to the task of symbolic discovery of SDEs~(Fig.~\ref{fig:overview}b+c), and we showcase this approach to a range of system identification problems~(Fig.~\ref{fig:overview}d). Our results show that the stochastic component can be learned effectively and improves drift recovery compared to evolving only an ODE. The method is competitive with the Kramers–Moyal–based sparse regression on low-dimensional problems, while demonstrating better scalability to higher-dimensional systems. Furthermore, incorporating integration enhances equation recovery from sparsely sampled data. Finally, we show that the approach can also identify the structure of SPDEs using the MLE. The equations discovered by the framework provide knowledge about the underlying systems, but also produce generative samples. Overall, the proposed approach provides a scalable and effective alternative for identification of stochastic systems, an important direction in scientific discovery.

\section{Related work}
Since symbolic regression was introduced as a possible application of GP~\cite{koza1994genetic}, it has gained much attraction in the domain of system identification~\cite{makke2024interpretable, angelis2023artificial}. Many tools for symbolic regression have been developed over the years, either based on GP~\cite{bongard2007automated, schmidt2009distilling, cranmer2023interpretable}, sparse regression~\cite{bongard2007automated} or physics-informed approaches~\cite{udrescu2020ai}, and several benchmarks were introduced~\cite{udrescu2020ai, d2023odeformer, la2021contemporary}. Symbolic regression is a popular method for discovering symbolic equations of physical and natural laws, including gene regulatory networks~\cite{sakamoto2001inferring}, energy consumption~\cite{castelli2015prediction} or chemical engineering~\cite{mckay1997steady}.

ODEs form a prominent class of systems to which symbolic regression has been widely applied~\cite{bongard2007automated, brunton2016discovering, d2023odeformer, quade2016prediction, cao2000evolutionary, iba2008inference}. From observed time-series data, symbolic expressions for the governing ODEs can be inferred, yielding interpretable models of the underlying dynamics. Traditionally, candidate equations are evaluated by numerically integrating the ODE system and comparing the resulting trajectories to the data, which can be computationally expensive. Instead of numerically integrating a system of ODEs, the finite difference method turns the evaluation into a regression task that maps system states to their derivatives for each variable independently. This substantially reduces the computational costs and simplifies the problem. SINDy and its extensions are based on the finite differences method~\cite{brunton2016discovering}, and it has also been adopted as a fitness evaluation method in GP~\cite{bongard2007automated, cranmer2023interpretable}.

Since SINDy has demonstrated the possibility of using sparse regression for symbolic discovery of dynamical systems, it has been extended to various applications, including PDEs~\cite{messenger2021weak, zhang2023deep, reinbold2019data,}, partially observed systems~\cite{bakarji2023discovering, conti2026veni}, SDEs~\cite{boninsegna2018sparse, wang2022data, rajabzadeh2016robust, li2021data, li2025evolutionary, callaham2021nonlinear, huang2022sparse, jacobs2023hypersindy, nabeel2025discovering} and SPDEs~\cite{mathpati2024discovering}. All current methods for symbolic regression of SDEs replace the finite differences method with the Kramers-Moyal expansion~\cite{risken1989fokker}, which gives estimates of the coefficients of the drift and diffusion at every time step of the dynamics. The conventional approach bins the data before computing the coefficients, however using the raw time series introduces a bias-variance trade-off for low sampling frequencies and increases the computational costs~\cite{nabeel2025discovering}. After performing the Kramers-Moyal expansion, identification of the drift and diffusion can be treated as separate regression problems, for which SINDy is a strong candidate. While the combination of sparse regression and the Kramers–Moyal expansion is effective, it has several limitations. Accurate estimation of drift and diffusion coefficients requires high-frequency sampling and sufficient data per bin, making the method highly susceptible to the curse of dimensionality~\cite{honisch2011estimation}. Secondly, the combination functions as a two-stage approach, meaning imprecisions can occur at multiple stages during identification. A general limitation of sparse regression is the over-reliance on the selected terms in the library, resulting in incorrect recovery when the required basis functions are missing.

Symbolic regression of SDEs is a relatively unexplored field, but black-box approaches to fit models to SDEs from discrete observations have been developed in recent years. For example, neural or latent SDEs fit parameterized drift and diffusion functions to observed data and act as generative models for new sample trajectories using variational inference~\cite{elgazzar2024generative, tzen2019neural, jia2019neural}. Differently, physics-informed neural networks also demonstrate the ability to learn stochastic processes~\cite{yang2020physics, zhong2023pi, shin2023physics}. Both neural SDEs and physics-informed neural networks result in a black-box model with little interpretability, where GP can evolve symbolic expressions for such problems. When the structure of SDEs is fixed, parameter estimation can be done with Bayesian inference~\cite{sarkka2015posterior, batz2018approximate} or by minimizing the MLE~\cite{nielsen2000parameter, pedersen1995new, kristensen2004parameter}, and even parameter-less approaches exist~\cite{ait2002maximum}. However, these methods depend on a predefined model structure, which introduces biases and reduces the modelling capacity compared to searching for the best equation with symbolic regression. Furthermore, when the intervals between observations become substantially large, the MLE can be optimized more accurately by integrating proposed solutions multiple steps between observations~\cite{pedersen1995new, nowman1997gaussian}, improving performance on sparsely sampled data. 

GP has not been directly applied to learn the structure of SDEs, let alone SPDEs, but in Ref.~\cite{li2025evolutionary} the library of base functions for sparse regression is built with GP. Symbolic regression via GP makes the approach less reliant on the given function library than sparse regression generally is. Learning the drift and diffusion separately does not guarantee a consistent or valid model, whereas optimizing the MLE with GP enables their joint evaluation for each variable, yielding more coherent solutions. Furthermore, integrating the MLE as objective function removes the reliance on the Kramers-Moyal expansion with binning, allowing for better scalability to higher-dimensional problems.

\section{Methods}
\subsection{Stochastic differential equations}
SDEs extend ODEs by modelling the randomness in dynamical systems. In this paper, the general, time-invariant, form of the SDE is considered, defined as
\begin{equation}
    \dd \mathbf{x}(t) = f(\mathbf{x}(t))\dd t + G(\mathbf{x}(t))\dd \mathbf{W}.\label{eq:sde}
\end{equation}
Here, $\mathbf{x}(t) \in \mathbb{R}^N$ governs the $N$-dimensional state, $f(\mathbf{x}(t)) \in \mathbb{R}^N$ and $G(\mathbf{x}(t)) \in \mathbb{R}^{N\times N}$ represent the drift and diffusion respectively and $\mathbf{W} \in \mathbb{R}^{N}$ denotes a standard Wiener process. When $G(\mathbf{x}(t))$ depends on $\mathbf{x}$, the noise is called \textit{multiplicative} and a constant $G(\mathbf{x}(t))$ is called \textit{additive}. We assume that every variable in the system is driven by independent noise, therefore we can explicitly decompose the dynamics as
\begin{equation}
    f(\mathbf{x}(t)) = \begin{bmatrix}
        f_1(\mathbf{x}(t))\\
        \vdots\\
        f_N(\mathbf{x}(t))
    \end{bmatrix}, \quad G(\mathbf{x}(t))=\text{diag}(g_1(\mathbf{x}(t)),\dots,g_N(\mathbf{x}(t))),
\end{equation} where both $f_i(\mathbf{x}(t))$ and $g_i(\mathbf{x}(t))$ output scalars for $i=1,\dots,N$.

Extending this framework to stochastic partial differential equations (SPDEs) replaces the finite-dimensional state $\mathbf{x}(t)$ with a space-dependent field $u(t,\mathbf{x})$ defined on a spatial domain $\Omega\subset\mathbb{R}^d$. Concretely, the SPDE can be written as
\begin{equation}
    \dd u(t,\mathbf{x}) = f(u(t,\mathbf{x}))\dd t + g(u(t,\mathbf{x}))\dd W(t,\mathbf{x}), \quad\mathbf{x}\in\Omega.
\end{equation}
Here, $f$ and $g$ are defined in the same way as in Eq.~\eqref{eq:sde}, and $\mathbf{x}$ are the spatial coordinates. To numerically simulate the SPDE, $\mathbf{x}$ is discretized into evenly spaced points and the initial condition $u(0,\mathbf{x})$ and boundary conditions $u(t,0)$ have to be defined. 

In practical applications, time series are observed from a system with unknown dynamics, but to measure accuracy of the learned models, we will focus on simulated systems with known ground truths. Note that we assume full observability of the system, therefore all dimensions of the data are available and observation noise is not added. To generate a set of time series, the SDE is integrated over the time window [$t_0, T$], given an initial condition $\mathbf{x}(0)$ and a Wiener process $\mathbf{W}$. The integration of SDEs is performed with Diffrax~\cite{kidger2022neural}, using the stochastic Runge-Kutta method for Stratonovich SDEs~\cite{foster2024high} with time step $\Delta t=0.001$. Subsequently, observations are sampled from the integration at discrete times $t_k=t_0+k\tau$ with $k=0,\dots,K$ and $K=\flatfrac{T}{\tau}$, resulting in $\{\mathbf{x}(t_k)\}^K_{k=0}$. To gather a batch of time series, different initial conditions and Wiener processes are sampled and used in integration of the system.

\subsection{Genetic programming}\label{sec:GP}

\begin{figure}[t]\centering\begin{minipage}{13cm}
\begin{small}
\begin{algorithm}[H]
\textbf{Input} Number of generations $g$, population size $p$, fitness function $\mathcal{F}$
\begin{algorithmic}[1]
    \State population $P$ $\longleftarrow$ \{\}
        \For{$i$ in $p$}
            \State sample drift $\hat{f}$ and diffusion $\hat{g}$
            \State append $(\hat{f}, \hat{g})$ to $P$
        \EndFor
    \For{$j$ in $g$}
        \State compute fitness $F$ of $p$ on $\mathcal{F}$
         \State offspring $O$ $\longleftarrow$ \{\}
         \State determine ranks $R$ of $P$ with NSGA-II
         \State append Pareto front of $P$ to $O$
         \While{size($O$) $<$ $p$}
         \State randomly select parents $p_1, p_2$ from $P$ with tournament selection given $R$
            \State randomly select {\em reproduce} from \{crossover, mutation\}
            \State child $c$ $\longleftarrow$ {\em reproduce} $(\hat{f}, \hat{g})$ from $p_1$ and $p_2$
        \State append $c$ to $O$
        \EndWhile
        \State $P$ $\longleftarrow$ $O$
    \EndFor
    \State \Return fittest individual in $P$
\end{algorithmic}
\caption{Genetic programming algorithm}\label{alg: GP}
\end{algorithm}\end{small}\end{minipage}\end{figure}

GP is an evolutionary algorithm originally introduced to optimize the structure of computer programs~\cite{koza1994genetic} and has since become a powerful method for symbolic regression. GP evolves both the structure and parameters of mathematical expressions to fit data with a population-based, gradient-free optimization algorithm utilizing mechanisms inspired from biological evolution. Starting from a randomly initialized population of $p$ individuals with maximum depth $m$, GP iteratively performs evaluation, selection, and reproduction over $g$ generations. The GP algorithm, adjusted to evolve the drift and diffusion, is presented in Algorithm~\ref{alg: GP}. In this paper, we made use of the Kozax library~\cite{de2025kozax}, a fast and flexible genetic programming framework built in JAX.

Individuals are represented as parse trees constructed from a predefined library of function and leaf nodes. An example tree is shown in (Fig.~\ref{fig:overview}a), which corresponds to the symbolic equation $\sin(x_1) \times (3 - x_2)$. Candidate solutions are assessed with a fitness function $\mathcal{F}$ that assigns a fitness value. After evaluating the population, individuals are selected for reproduction, favouring higher fitness. In Kozax, solutions are selected following the NSGA-II algorithm~\cite{deb2002fast} with fitness and complexity as objectives, defined as the number of nodes. The selected individuals undergo structural changes during reproduction with crossover and mutation. Crossover swaps randomly selected subtrees between individuals (Fig.~\ref{fig:overview}b), while mutation modifies operators or leaves, inserts or deletes nodes (Fig.~\ref{fig:overview}c). During reproduction the number of nodes in a tree is constrained by the hyperparameter $s$.

To evolve the drift and diffusion simultaneously, individuals consist of multiple trees~\cite{langdon1998genetic}, where the computed fitness corresponds to the combination of the trees. During reproduction, crossover can only be applied to a pair of trees at the same position in two individuals. Furthermore, a whole-tree crossover operation is included to foster the exchange of good trees~\cite{muni2004novel}. To traverse the solution space more efficiently in the enlarged solution space, the constants in trees are optimized with gradient descent~\cite{topchy2001faster}. To limit computational costs, this optimization is applied to the subset of $p^*$ individuals, prioritizing the fittest solutions, for $g^*$ iterations using step size $\eta$. To prevent premature convergence and preserve diversity, the population is divided into multiple subpopulations~\cite{fernandez2003empirical}, each with distinct probability for crossover and mutation~\cite{herrera2000gradual}. Every couple of generation, individuals are migrated to introduce diverse and promising structures in other subpopulations~\cite{fernandez2003empirical}. The hyperparameters that were used in the implementation of the GP algorithm are presented in Appendix~\ref{sec:App_hyp}, Table~\ref{Table: Hyper_params_GP}.

\subsection{Problem definition}
The central task in this paper is identifying the governing equations from observed noisy time series data~(Fig.~\ref{fig:overview}a). We introduce a method to evolve symbolic equations of SDEs with GP. The effectiveness of our method is investigated in a set of experiments, and is compared to evolving ODEs and sparse regression of the Kramers-Moyal expansion.

\subsubsection{Experiments}
To demonstrate the effectiveness and robustness of our method, several systems and settings are investigated. First, the one-dimensional double well problem is identified with additive, linear multiplicative and non-linear multiplicative noise to test for accurate recovery given increasingly complex diffusion functions. Following, we study the van der Pol oscillator and Rössler attractor, both extended with multiplicative noise. To test for scalability to high-dimensional problems, the Lorenz96 model with five, ten and twenty variables is subjected to symbolic regression. The Lotka-Volterra model is studied with varying sampling rates to examine performance given limited data. Fig.~\ref{fig:overview}d shows example trajectories of these environments. Finally, the Fisher-KPP and two dimensional heat equations are studied to test for generalization to SPDEs. For more details about each environment, we refer to Appendix~\ref{sec:App_exp}. In all experiments, the node set in GP consists of $+$, $\times$ and the number of variables in an environment. In case of the SPDEs, the gradients and Laplacian are added to the node set too.

\subsubsection{Fitness function}
GP jointly evolves symbolic expressions for the component-wise drift $\hat{f}_i(\mathbf{x}(t))$ and diffusion $\hat{g}_i(\mathbf{x}(t))$ functions for a variable $x_i$. To evaluate the fitness of a candidate solution, the MLE is determined given data. Our approach focuses on normally distributed noise, therefore the MLE can be defined as the negative log likelihood of a Gaussian distribution. The MLE is computed for each state transition, representing the probability of $x_i(t)$ given the mean $\mu_i(t)$ and variance $\sigma_i^2(t)$ at time $t$ for $i=1,...,N$. The fitness is calculated by summing the pairwise conditional transition probabilities, similar to the simulation-free finite differences method~\cite{brunton2016discovering}. Since full observability is assumed, the drift and diffusion can be optimized separately for each of the variables in the target system. This reduces the search space, as fewer trees have to be evolved simultaneously. We will refer to this approach as GP-SDE in the results section. Concretely, the fitness function for a pair of functions is defined as
\begin{equation}
    F(\hat{f}_i,\hat{g}_i) \coloneq \log p(\{x_i(t_k)\}^K_{k=0} \mid \{\mu_i(t_k)\}^K_{k=0}, \{\sigma^2_i(t_k)\}^K_{k=0}) = \sum_{k=1}^{K} 
    \left[ \frac{1}{2}\log(2\pi\sigma_i(t_k)^2) + \frac{(x_i(t_k) - \mu_i(t_k))^2}{2\sigma_i(t_k)^2}\right]
\end{equation}
with $\mu_i(t_k) = x_i(t_{k-1}) + \tau \hat{f}_i(\mathbf{x}(t_{k-1}))$ and $\sigma(t_k) = \sqrt{\tau} \hat{g}_i(\mathbf{x}(t_{k-1}))$.

When the sampling rate of the observed data is too low, directly minimizing the MLE of transition pairs could result in inaccurate models. Alternatively, equations for multiple variables can be evolved together~\cite{cao2000evolutionary}. Hence, this allows for numerical integration of the system of equations between the observations with a smaller time step~\cite{pedersen1995new, nowman1997gaussian}. Subsequently, the mean of the next state is computed by $\boldsymbol{\mu}(t_k) = \mathbf{y}(L)$ with $\mathbf{y}(l)=\mathbf{y}(l-1) + \frac{\tau}{L} \hat{f}(\mathbf{y}(l-1))$ and $\mathbf{y}(0) = \mathbf{x}(t_{k-1})$ for $l=0,...,L$, where L is set to 5. Similarly, the variance changes to $\boldsymbol{\sigma}(t_k)=\frac{\sqrt{\tau}}{L}\sum_{l=0}^{L-1} \hat{g}(\mathbf{y}(l))$. The extension of GP-SDE with multiple steps (MS) of integration is referred to as GP-SDE-MS in the remainder of the results section. 

\subsubsection{Baselines}\label{sec:baselines}
\paragraph{Ordinary differential equations} Symbolic regression of deterministic dynamical systems has shown great results, even recovering the correct ODE equations from noisy observations~\cite{d2023odeformer, reinbold2021robust}. However, the performance of such methods may deteriorate when the dynamics are stochastic. In this case, it will be beneficial when the stochastic component can be recovered with symbolic regression as well. To demonstrate the advantages of our method, we compare with the conventional GP approach that only learns a drift term $\hat{f}_i$ for each variable from data, which we refer to as GP-ODE. The fitness function that will be minimized with GP-ODE is 
\begin{equation}
    F(\hat{f}_i) \coloneq \text{MSE}(\{x_i(t_k)\}^K_{k=0} \mid \{\mu_i(t_k)\}^K_{k=0}) = \sum_{k=1}^{K} (x_i(t_k) - \mu_i(t_k))^2,
\end{equation}
where $\mu_i(t_k) = x_i(t_{k-1}) + \tau \hat{f}_i(\mathbf{x}(t_{k-1}))$. GP-ODE can also be extended to integrate the drift over multiple steps between observations, which will be called GP-ODE-MS.

Furthermore, our proposed method will also be compared to established work for symbolic recovery of SDEs, based on the Kramers-Moyal expansion and sparse regression, which we will name KM-SR. The Kramers-Moyal expansion and sparse regression will be described shortly, but refer to Ref.~\cite{boninsegna2018sparse} for a more detailed explanation.

\paragraph{Kramers-Moyal expansion}
The Kramers-Moyal expansion is a Taylor expansion of time series data that gives explicit estimates of the drift $D_{i,1}$ and diffusion $D_{i,2}$ coefficients of variable $x_i$. The coefficients can be estimated with
\begin{align} 
    D_{i,1}(\{x_i(t_k)\}^K_{k=0}) &= \mathbb{E}_{k \in \left[0,K\right]} \left[ \frac{1}{\tau} (x_i(t_k) - x_i(t_{k-1})) \right], \quad
    D_{i,2}(\{x_i(t_k)\}^K_{k=0}) = \sqrt{ \mathbb{E}_{k \in \left[0,K\right]} \left[ \frac{1}{\tau} (x_i(t_k) - x_i(t_{k-1}))^2 \right]}.\label{eq: KM2}
\end{align}
To improve computational efficiency and accuracy of the estimates, the coefficients of the drift and diffusion are averaged within bins. The bins are determined by uniformly dividing the range between the lowest and highest value from the observed data into a chosen number of bins $b$. For multi-dimensional systems this results in a grid, where each dimension is divided into $b$ bins. The coefficients of a bin are only included if the number of data points within that bin is large than a specified threshold $\beta$.

\paragraph{Sparse regression}
Sparse regression combines L1-regularization on regression with a function library to learn symbolic equations from data, and it became a strong alternative to GP for symbolic regression~\cite{brunton2016discovering}. The function library $\boldsymbol{\Theta}$ is constructed by computing all combinatorial polynomials of the observed variables up until a specified degree $d$. The function library is multiplied with a vector $\mathbf{c}$, obtained by fitting it to the observed data through least-squares regression with L1-regularization. The optimal coefficient vector $\mathbf{c}^*$ is determined by
\begin{equation}\label{Eq: sparse}
    \mathbf{c}^* = \argmin_{\mathbf{c}} \biggl[\lVert \mathbf{y} - \mathbf{c}\boldsymbol{\Theta} \rVert_2 + \alpha\lVert \mathbf{c} \rVert_1 \biggr].
\end{equation}
Here, $\alpha$ determines the strength of the L1-regularization, with a higher alpha resulting in sparser solutions. After finding $\mathbf{c}^*$, the coefficients lower than threshold $\lambda$ are set to zero and the corresponding functions are removed from the library to get $\boldsymbol{\Theta}'$~\cite{callaham2021nonlinear}. Equation~\eqref{Eq: sparse} is run again with $\boldsymbol{\Theta}'$, repeating until no coefficients in $\mathbf{c}^*$ are smaller than $\lambda$.

The Kramers-Moyal expansion returns the coefficients of the drift and diffusion of one variable of the observed time series, after which sparse regression is applied separately for the drift and diffusion, resulting in the two equations. For systems with multiple dimensions, the Kramers-Moyal expansion is performed independently for each variable. Both stages in KM-SR are sensitive to the selected hyperparameter values, and needs tuning for different problem settings. To improve generalizability of KM-SR, a hyperparameter search is performed for the drift and diffusion independently for each variable in every experiment. The hyperparameters that can be tuned are $b$, $\beta$, $\alpha$ and $\lambda$. The sets of values for each hyperparameter in the different experiments can be found in Table~\ref{Table: Hyper_params_KMSR}.

\paragraph{Comparison}
The three methods, KM-SR, GP-ODE and GP-SDE are each optimized with different metrics, therefore direct comparison is not trivial. To evaluate the methods fairly, three datasets are generated: an optimization set, a validation set and a test set. The optimization set is used to evolve or fit models to match the data. The true system is applied to the validation and test sets, resulting in the target values for the drift and diffusion. This allows for computing the mean squared error (MSE) between estimated and target outputs of the drift with
\begin{equation}
    \text{MSE}(\{\mathbf{x}(t_k)\}^K_{k=0}) = \frac{1}{NK} \sum_{i=1}^N \sum_{k=0}^K \left[ f_i(\boldsymbol{x}(t_k)) - \hat{f}_i(\boldsymbol{x}(t_k))\right]^2,
\end{equation}
where $f_i(\mathbf{x}(t))$ is the ground truth drift function and $\hat{f}_i(\mathbf{x}(t))$ is the discovered drift function with any method. Similarly, the MSE of the diffusion is computed with $g_i(\mathbf{x}(t))$ and $\hat{g}_i(\mathbf{x}(t))$. KM-SR produces different solutions during the hyperparameter search, and the solution with the best MSE on the validation set is returned. In GP-ODE and GP-SDE, the solution from the final Pareto front with the lowest validation MSE is selected. Finally, the MSE on the test data is computed for the best solution of each method for comparison.

\subsubsection{Hardware setup}
The GP-based methods were performed on an Intel Xeon Platinum 8360Y (2.40 GHz) 36-core machine with a single NVIDIA A100 GPU (40 GiB HBM2 memory), while KM-SR ran on an AMD Genoa 9654 (2.40 GHz) with 192 cores. 

\section{Results}
In this paper, we introduce a method based on genetic programming for identifying stochastic differential equations (GP-SDE). GP-SDE is compared to a standard approach in symbolic regression, evolving only the drift function with genetic programming (GP-ODE). Furthermore we compare to the established baseline based on the Kramers-Moyal expansion and sparse regression (KM-SR). The methods are evaluated on various environments, investigating recovery of systems given different noise terms, dimensionality and sampling rates. The code and data can be found at \url{https://github.com/sdevries0/GP-SDE}.

\begin{figure}[t]
    \centering
    \includegraphics[width=\linewidth]{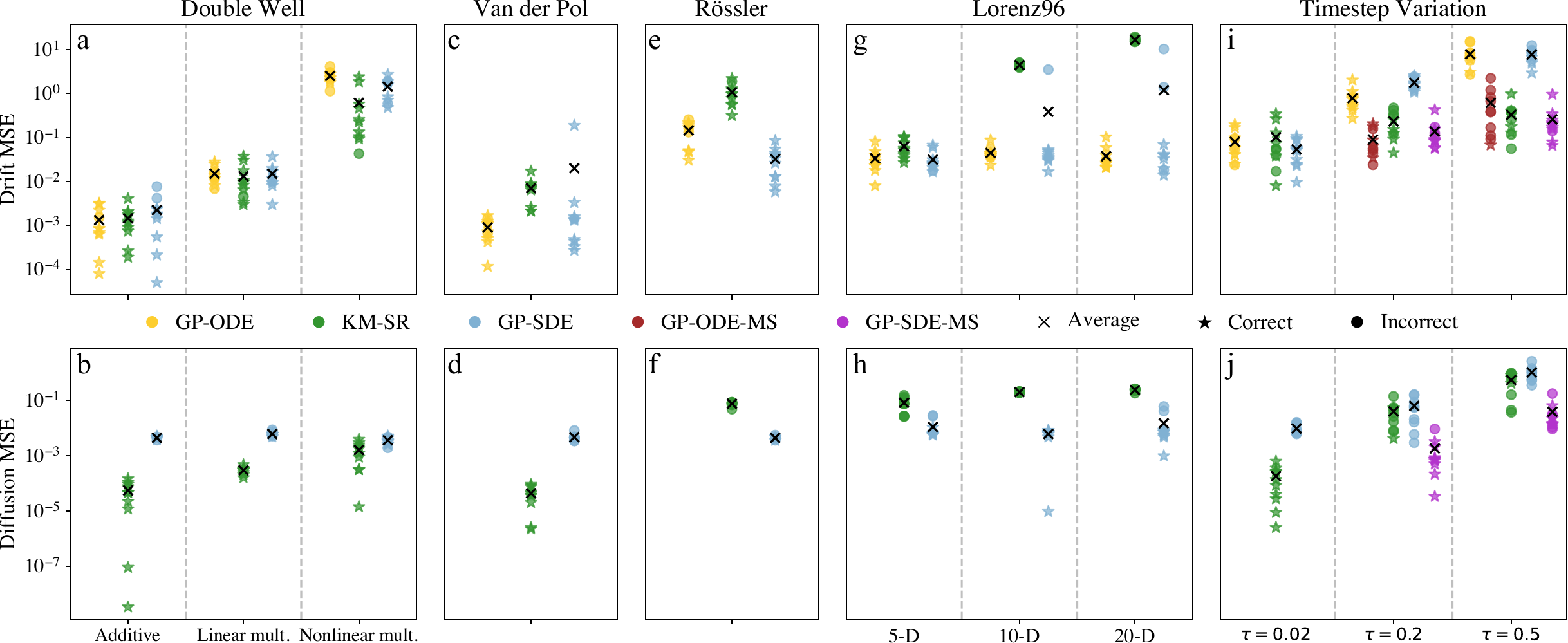}
    \caption{\textbf{GP-SDE accurately recovers the governing equations of stochastic dynamical systems.} The mean squared error (MSE) between the true and learned functions for the drift and diffusion on test data are presented for ten different seeds. The average of the ten runs in indicated with a black cross for every method. Furthermore, for each run, a star indicates that the correct structure of the equations were found for all variables, while a circle indicates that the structure of at least one equation was incorrect. The methods include Kramers-Moyal expansion followed by sparse regression (KM-SR), genetic programming of ordinary differential equations (GP-ODE), which only learns a function for the drift, and genetic programming of stochastic differential equations (GP-SDE). (a, b) Results for the double well problem with an additive, linear and non-linear multiplicative diffusion term. (c, d) Results for the van der Pol oscillator. (e, f) Results for the Rössler attractor. (g, h) Results for the Lorenz96 model with 5, 10 and 20 variables. (i, j) Results for the Lotka-Volterra model with a sampling rate of 0.02, 0.2 and 0.5. Both GP-ODE and GP-SDE are extended with multi-step integration, labelled as GP-ODE-MS and GP-SDE-MS respectively.}
    \label{fig:results}
\end{figure}

\subsubsection*{GP-SDE is competitive with KM-SR on low-dimensional problems}
To assess whether GP-SDE can accurately recover the drift and diffusion from data, we first study the one-dimensional double well problem with different diffusion terms. The MSE of the drift and diffusion obtained with GP-ODE, KM-SR and GP-SDE are presented in Figs.~\ref{fig:results}a+b. For the additive (scalar) diffusion, GP-ODE consistently recovered the correct drift function with low MSE. KM-SR achieves comparable drift accuracy, while also identifying the diffusion. GP-SDE performed competitively, albeit less consistently across different runs. 

Changing the diffusion in the double well problem to multiplicative complicates the task due to more irregular system dynamics. With linear multiplicative noise, the MSE of GP-ODE increased, indicating sensitivity to the process noise. KM-SR still recovered the correct structures for both drift and diffusion, thought with higher MSE. GP-SDE exhibits similar performance as KM-SR, but slightly higher diffusion MSE. With non-linear multiplicative noise the scale of the MSE drastically increased with each method. GP-ODE is consistently outperformed by the other approaches, demonstrating that accurate drift recovery is not possible without modelling the complex diffusion term. KM-SR generally outperforms GP-ODE, but shows degraded performance for some seeds. Again, KM-SR is slightly more accurate than GP-SDE in terms of MSE, but both methods often evolved the correct equations.

The experiments on the double well problem demonstrated that GP-SDE is a competitive method to KM-SR in one-dimensional settings, although KM-SR is more precise, and evolving the diffusion term has advantages over GP-ODE. To further investigate the strength and limitations of each method, we consider benchmark environments with stochastic multi-dimensional dynamics. The first benchmark environment was the stochastic van der Pol oscillator (Figs.~\ref{fig:results}c+d), where the diffusion term adds multiplicative noise to only one state variable, requiring the methods to distinguish between stochastic and deterministic variables. The drift is identified most accurately with GP-ODE, while KM-SR correctly identifies the structure of both drift and diffusion. GP-SDE achieved comparable drift MSE to the baselines, although one run resulted in an incorrect equation. The structure of the diffusion is not recovered in any run with GP-SDE, but still the MSE remains relatively close to KM-SR.

The Rössler attractor further complicates the task, as it is a three dimensional system with chaotic behaviour, (Figs.~\ref{fig:results}e+f). Still, GP-ODE manages to recover the correct equation in some runs. The higher dimensionality and chaotic trajectories reduce the accuracy of the binning in KM-SR, resulting in degraded MSE for both the drift and diffusion, despite often finding the correct structure. In contrast, GP-based methods do not rely on binning, enabling GP-SDE to outperform KM-SR in MSE for both components. Moreover, GP-SDE benefits from also learning the diffusion term, as GP-SDE achieves lower average drift MSE and more consistent recovery than GP-ODE.

\subsubsection*{GP-SDE remains robust in high-dimensional problems}
The Rössler attractor highlights that increasing dimensionality reduces the robustness of KM-SR due to inaccurate binning and a growing function library. To further investigate this limitation, the methods are evaluated on the stochastic Lorenz96 model with five, ten and twenty variables, of which the MSE is presented in Fig.~\ref{fig:results}g+h. The correct drift functions were recovered with GP-ODE for all three cases. With five variables, KM-SR still performs well, achieving competitive MSE, but its performance degrades substantially for ten and twenty variables, failing to accurately identify both drift and diffusion. This degradation is explained by poor scalability of the binning procedure. With ten variables, using four bins in each dimension already exceeds a million bins in total and the size of the function library is 66, which increases even further with twenty variables. GP-SDE scales more favourably than KM-SR, as lower MSEs were achieved in all settings and even matched GP-ODE’s drift accuracy in the five-dimensional case. Although GP-SDE does not consistently recover the exact equations for ten and twenty variables, it evolves accurate solutions in most runs and clearly outperforms KM-SR.

\subsubsection*{With integration GP-SDE can identify equations from sparse data}
In the previous experiments the trajectories were densely sampled, making equation recovery relatively easy. However, in practice the data might be sampled sparsely or irregularly. To evaluate robustness under sparse data conditions, the methods are compared on the stochastic Lotka-Volterra model with varying sampling rates (Figs.~\ref{fig:results}i+j). At the original sampling time of 0.02, KM-SR, GP-ODE and GP-SDE achieve competitive performance, where KM-SR is more accurate for diffusion, while GP-SDE yields better drift recovery. Increasing the sampling time to 0.2 led to a slight rise of the MSE for KM-SR, while GP-ODE and GP-SDE performed substantially worse, especially for the drift term. Since GP enables simultaneous evolution of the drift and diffusion for multiple variables, the functions can be numerically integrated between observed data points to get more fine-grained predictions. With this addition, GP-ODE-MS and GP-SDE-MS improved over their standard counterparts, with both methods outperforming KM-SR. GP-SDE-MS is also the most successful at identifying the correct equation structures. At a sampling time of 0.5, the benefit of multi-step integration becomes even more apparent. Here, GP-SDE-MS outperforms GP-ODE-MS, further demonstrating that modelling stochasticity improves the recovery of deterministic functions.

\subsubsection*{The learned equations produce informative generative samples}
Fig.~\ref{fig:results}e and f show that GP-SDE outperforms KM-SR and GP-ODE on the Rössler attractor in terms of MSE. All three methods are able to recover the correct structure of the drift in most runs, while the diffusion structure is not identified with both KM-SR and GP-SDE. To inspect the learned systems, Table~\ref{Table: Rossler_eq} presents the equations of the best run with each method, as well as the ground truth equations. GP-ODE recovered the correct equation and parameters for $f_x$ and $f_y$, and the recovered $f_z$ has minor deviations in the constants. The major limitation is that no functions are learned to capture the stochasticity. While KM-SR does identify a function for the diffusion, the fitted equations resemble the ground truth less well, even though the correct equation structures were identified for the drift. KM-SR missed to include multiplication with the state in $g_x$, and includes an additional constant to $g_z$. GP-SDE combines both learning the drift and diffusion with highly accurate recovery. The constants in the drift are close to the ground truth, and the fit is better than GP-ODE for $f_z$. The diffusion terms are close to the true functions, albeit slightly lower constants. Similar to KM-SR, GP-SDE also learned an additional constant in $g_z$, but with a value closer to zero, while also achieving a better fit on the scalar of the state.

\begin{table}[ht]
\centering
\caption{The equations corresponding to the true system and best solutions discovered of the Rössler attractor with genetic programming for ordinary differential equations (GP-ODE), Kramers-Moyal expansion and sparse regression (KM-SR) and genetic programming for stochastic differential equations (GP-SDE). For each variable $x$, $y$ and $z$, the equation for the drift and diffusion are shown. GP-ODE only evolves a function for the drift, therefore no diffusion equations are shown.}\label{Table: Rossler_eq}\vspace{0.2cm}
\begin{small}\centering
\begin{tabular}{lcccccc}
 \toprule
 \textbf{Method} & $f_x(x, y, z)$ & $g_x(x, y, z)$ & $f_y(x, y, z)$ & $g_y(x, y, z)$ & $f_z(x, y, z)$ & $g_z(x, y, z)$\\
 \midrule
 True & $-x - y$ & $0.100x$ & $x + 0.200y$ & $0.100y$ & $xz- 5.700z + 0.200$ & $0.100z$\\
 GP-ODE & $-0.999x - 0.999y$ & $-$ & $x + 0.196y$ & $-$ & $xz- 5.790z + 0.282$ & $-$\\
 KM-SR & $-0.950x - 1.486y$ & $0.306$ & $1.003x + 0.198y$ & $0.146y$ & $0.868xz- 4.849z + 0.157$ & $0.064z + 0.055$\\
 GP-SDE & $-x - 0.981y$ & $0.083x$ & $x + 0.191y$ & $0.083y$ & $xz- 5.670z + 0.188$ & $0.091z - 0.008$\\
 \bottomrule
\end{tabular}
\end{small}
\end{table}

\begin{figure}[!ht]
    \centering
    \includegraphics[width=\linewidth]{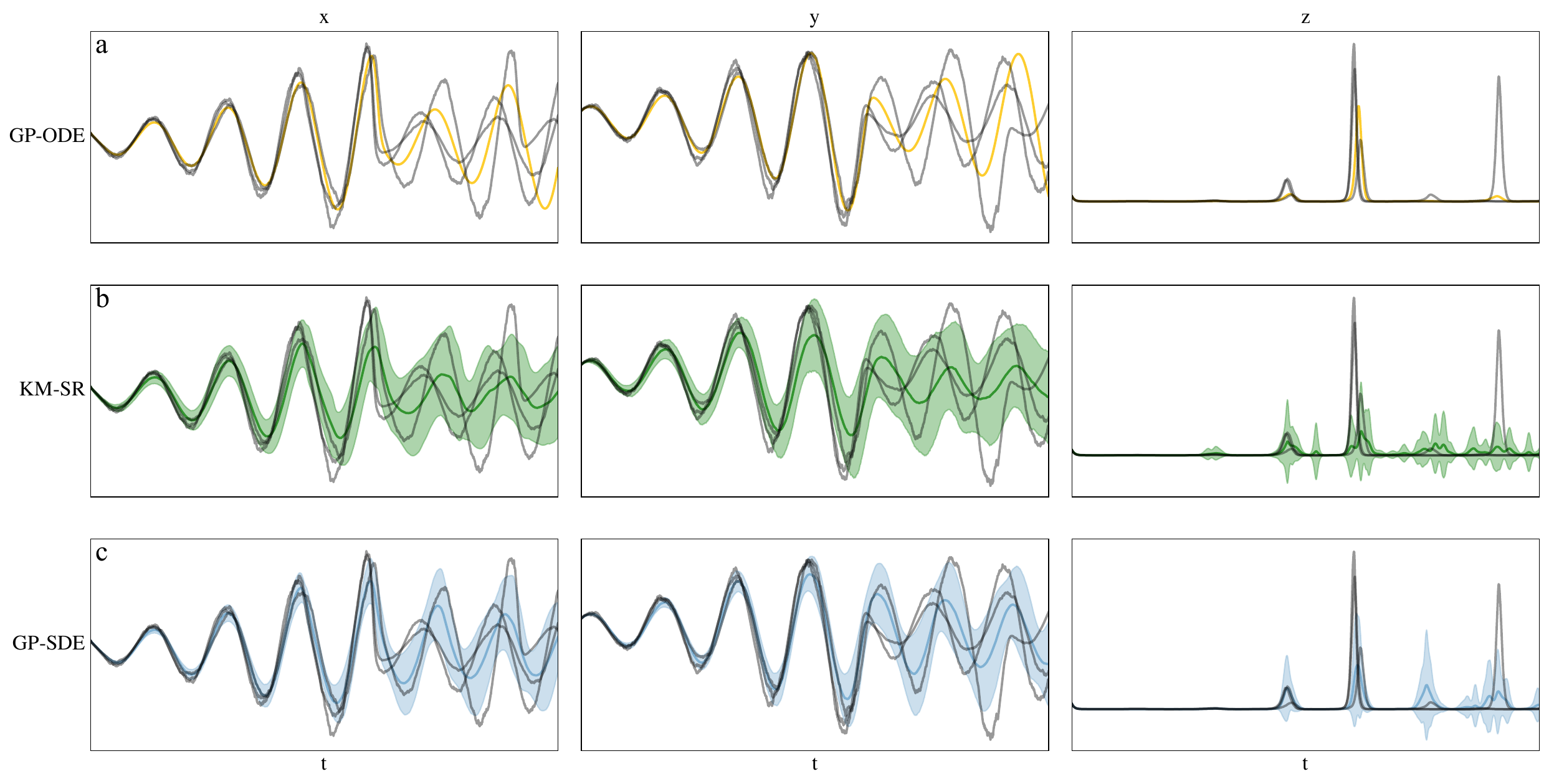}
    \caption{\textbf{Simulation of the discovered models for the Rössler attractor.} (a) Given a fixed initial condition, three trajectories of the Rössler attractor are simulated with different Wiener processes (black). The best ordinary differential equation evolved by genetic programming (GP-ODE) is integrated from the same initial condition (orange). (b) The same three trajectories of the true system are shown, together with the mean (green) and standard deviation (shaded area) computed over 50 trajectories sampled from the best model found by method based on the Kramers-Moyal expansion and sparse regression (KM-SR), where the initial condition was fixed but different Wiener processes were used in each trajectory. (c) Same approach as in (b), but with the best stochastic differential equation discovered with genetic programming (GP-SDE). For each method, the corresponding equations are presented in Table~\ref{Table: Rossler_eq}.}
    \label{fig:rossler_comparison}
\end{figure}

To further analyse the quality of the found equations, the systems are simulated and compared to true trajectories. First, we generate three trajectories with the ground truth given a fixed initial condition. The true time series are shown in black in Fig.~\ref{fig:rossler_comparison}a, where the deterministic trajectory of the system learned with GP-ODE is illustrated in yellow. Since GP-ODE returns a deterministic system, only a single trajectory can be obtained given an initial condition. This deterministic trajectory shows similar pattern as the true trajectory in $x$ and $y$, but diverges from the true samples after some time. The trajectory of $z$ includes the large upward spike at the right time point, but misses the minor spikes. Since only one sample can be generated from the deterministic system, the evolution of possible sample paths cannot be shown.

Since KM-SR also returns a diffusion term, generating multiple trajectories results in a coverage of possible paths of the SDEs. The average and standard deviation over 50 samples are presented in Fig.~\ref{fig:rossler_comparison}b in green. The spread of sampled paths of $x$ and $y$ already expands early in the trajectory. Eventually the samples cover a large part of the state space without closely resembling the true time series, particularly at the extreme values of the true trajectories. The samples of variable $z$ show spikes more frequently with inaccurate heights and at incorrect time points compared to the ground truth samples. 

The mean and standard deviation of the samples generated with the solution of GP-SDE are illustrated in blue in Fig.~\ref{fig:rossler_comparison}c. These samples show a better fit, with a more precise mean and standard deviation for variable $x$ and $y$, covering the extreme values and appearing more oscillatory. The spikes of variable $z$ largely overlap with the true trajectories, although incorrect timing and height also occurs. These simulations demonstrate that valid systems can be identified given the proposed setup, while showcasing the benefits of finding stochastic solutions and effectiveness of our approach over KM-SR.

\subsubsection*{GP-SDE extends to stochastic partial differential equations}
\begin{figure}[t]
    \centering
    \includegraphics[width=\linewidth]{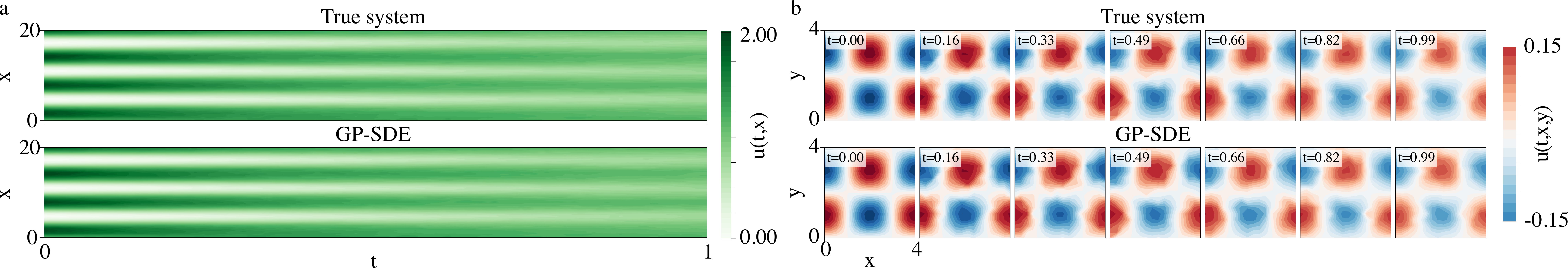}
    \caption{\textbf{GP-SDE can recover stochastic partial differential equations.} (a) Evolution of the Fisher-KPP equation with respect to $x$ and time, given the ground truth equation and identified system with GP-SDE, presented in Table~\ref{Table: SPDE_eq}. (b) Snapshots of the evolution of the two-dimensional heat transfer with respect to $x$, $y$ and time. Again, the true and learned system with GP-SDE are presented.}
    \label{fig:SPDE}
\end{figure}

To further test the generalization of the proposed methodology, GP-SDE was applied to learn the governing equations of SPDEs. First, the one-dimensional Fisher-KPP equation with multiplicative noise was investigated. GP-SDE recovered the correct structure for the drift with slightly inaccurate constants, and a small constant was added in the stochastic component (Table~\ref{Table: SPDE_eq}). Simulation of the evolution of the Fisher-KPP equation shows high similarity between the true and fitted systems (Fig.~\ref{fig:SPDE}a). To also test for scalability in the SPDE regime, GP-SDE was applied to the two-dimensional heat transfer system. Here, GP-SDE evolved the correct structure for the full equation, and the constants are very close to the ground truth (Table~\ref{Table: SPDE_eq}). As expected, the simulation of the true and evolved system show similar patterns. These experiments demonstrate that our approach correctly identifies the structure SPDEs, while still using the MLE as objective. 

\begin{table}[t]
\centering
\caption{The system corresponding to the true stochastic partial differential equations and the solutions evolved with GP-SDE are presented.}\label{Table: SPDE_eq}\vspace{0.2cm}
\begin{small}\centering
\begin{tabular}{lcc}
 \toprule
 \textbf{Method} & Fisher-KPP equation & 2-D heat transfer \\
 & $\dd u(t,x)$& $\dd u(t,x, y)$\\
 \midrule
 Ground truth & $\left(u_{xx} + u (1-u) \right)\dd t + 0.100 u \dd W$ & $ 0.100 (u_{xx} + u_{yy})\dd t + (u_x + u_y) \dd W$ \\
 GP-SDE & $\left( u_{xx} + u(0.929 - 0.936u) \right)\dd t + (0.094u - 0.011) \dd W$ & $ 0.098 (u_{xx} + u_{yy})\dd t + (u_x + u_y) \dd W$ \\
 \bottomrule
\end{tabular}
\end{small}
\end{table}

\subsubsection*{The runtime of GP-SDE scales efficiently} 
\begin{figure}
    \centering
    \includegraphics[width=\linewidth]{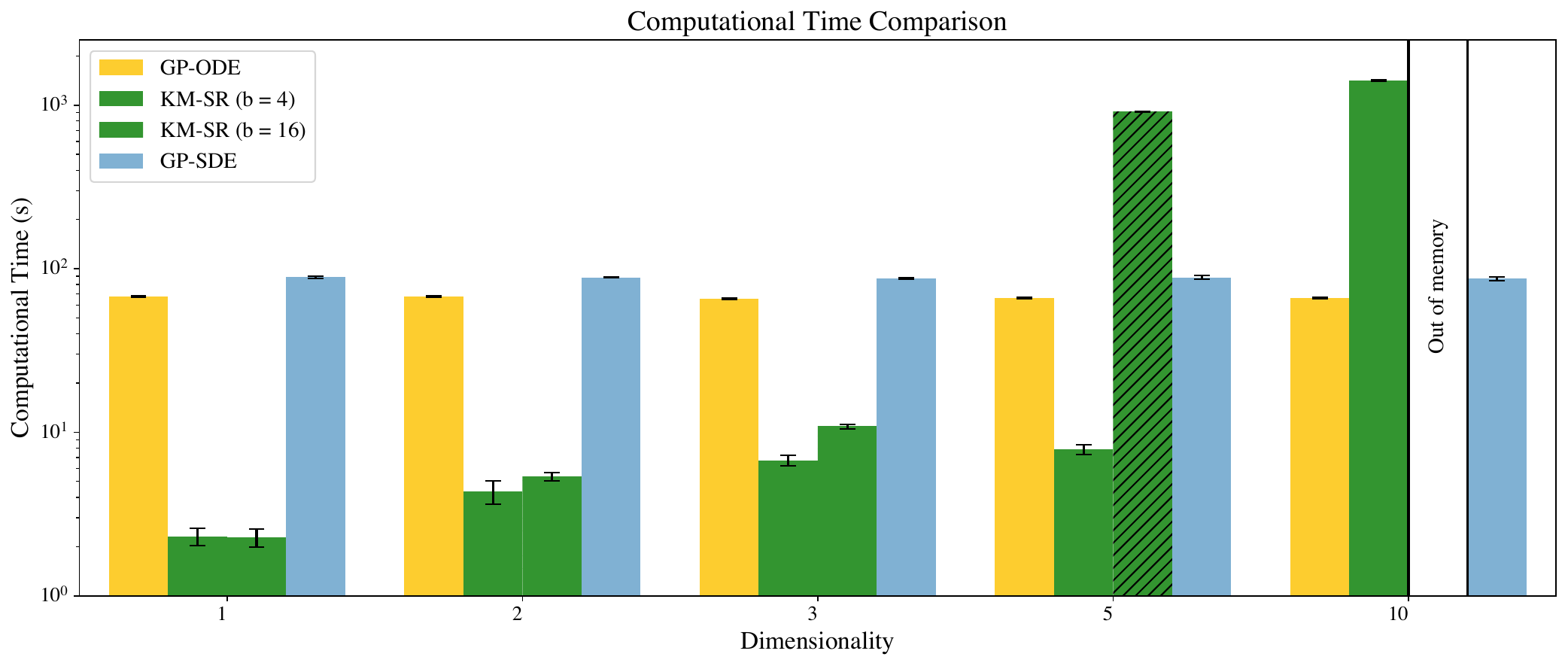}
    \caption{\textbf{Runtimes of the methods.} The computational time of GP-ODE, KM-SR and GP-SDE on the Lorenz96 model with increasing dimensionality. The runtimes are averaged over the ten seeds for every experiment, after running one seed to compile the algorithms. KM-SR is evaluated with four and sixteen bins ($b$).}
    \label{fig:time}
\end{figure}

The experiments have demonstrated that fitting SDEs with GP provides better scalability and robustness to sparse data constraints. However, for practical applications the computational time of the methods also is an important attribute. In Fig.~\ref{fig:time}, the average computational times of GP-ODE, KM-SR and GP-SDE for the Lorenz96 model with varying dimensionality are presented. The runtime of KM-SR is evaluated with four and sixteen bins, while still performing the hyperparameter search over the other parameters. GP-ODE and GP-SDE were evaluated with a population size of 500 for 50 generations. With low dimensionality, KM-SR is substantially faster than the GP-based method, where more bins slightly increased the runtime. However, with sixteen bins the runtime exploded when the dimensionality is five, and with a dimensionality of ten this configuration of KM-SR was computationally infeasible. Even with four bins, KM-SR became extremely slow on the ten dimensional problem, while also producing inaccurate results (Fig.~\ref{fig:results}g+h). Although GP-ODE and GP-SDE are slower than KM-SR for low-dimensional problems, the runtime barely increases with problem size. For all problems, GP-SDE is slightly slower than GP-ODE because more trees have to be evolved, but the relative difference does not enlarge with more variables. These results show that GP is a more sustainable approach as it scales well to higher dimensions, both computationally and effectively.

\section{Discussion}
In this paper, we introduced a method for data-driven discovery of symbolic equations of SDEs. A pair of equations for the drift and diffusion was learned with genetic programming by optimizing the MLE objective. Most methods for symbolic regression focus solely on learning a drift function, but identifying the diffusion both offers more modelling capacity and knowledge about the underlying system. Currently, the only alternative for symbolic regression of SDEs is a combination of the Kramers-Moyal expansion and sparse regression~\cite{boninsegna2018sparse}. Compared to this method, our introduced approach shows benefits such as that it does not rely on a two-step approach or a binning approach, is less sensitive to hyperparameters and the functions are optimized jointly, improving the validity of the resulting system with respect to the observed data.

We tested our method on various experiments, investigating benchmark systems, scalability and robustness to low sampling rates. Extending genetic programming to learn the diffusion demonstrated that both functions could be identified accurately, while not degrading the quality of the deterministic component. Furthermore, for more complex tasks, evolving both functions even proved to be beneficial for learning the drift function compared to only learning the drift. Genetic programming performed competitively to sparse regression of the Kramers-Moyal expansion, and even outperformed it in environments with higher dimensionality. The binning approach in the Kramers-Moyal expansion is very inefficient for many variables, especially if the range of data values is large too. Our method could recover the true equations even when the dimensionality increased to ten or twenty, while remaining considerably more computationally efficient. Furthermore, our approach could be extended to learn equations for all variables simultaneously, allowing for numerical integration between observed data points. With this extension, the equations were still recovered accurately from sparsely sampled data. Finally, the framework demonstrated to generalize to learn the structure from SPDEs as well.

A fundamental challenge in learning dynamical systems, and SDEs in particular, is identifiability and non-uniqueness~\cite{yamada1971uniqueness, browning2020identifiability}. Distinct systems can generate statistically indistinguishable dynamics, making it difficult to identify the true equations under sparse data constraints or noisy dynamics. In SDE inference, this challenge is enhanced by the explicit separation of the drift and diffusion. Stochastic effects may be partially absorbed by the drift, or deterministic structure are explained through the diffusion. While our results show that the correct function structures are often recovered, we occasionally observe accurate model fits were obtained with (partially) incorrect symbolic forms. Consequently, when applying this method to real-world systems, it is important to recognize that a good predictive fit does not necessarily imply that the correct governing equations were found.

Our approach relies on strong assumptions that may limit applicability to practical applications. Full observability of the data is assumed, but the observations could be noisy or variables may be unobserved. Adressing such settings requires learning latent SDEs using variational inference~\cite{elgazzar2024generative, tzen2019neural, hasan2021identifying}. SINDy has been extended to learn latent symbolic dynamics from partially observed data, however restricted the function space to deterministic systems~\cite{bakarji2023discovering, conti2026veni}. Extending symbolic regression to latent SDEs poses multiple challenges, including evolving the full system jointly, inferring initial conditions and integrating full trajectories. Moreover, the fitness function must be adapted to evaluate inferred latent trajectories given observed data. The issue of identifiability becomes even more pronounced, as separating the observation noise from the process stochasticity is inherently challenging.

Another assumption that could hinder practical usability is that the method is limited to identifying systems with normally distributed noise that is separable from the drift. Ref.~\cite{li2021data} showed that the KM-SR could also be used to identify terms that govern L\'evy jumps. The fitness function introduced in this paper should be generalized to learn equations for other types of noise, and ideally even select the most appropriate noise to explain the observed data. The approach could be generalized even further, removing the assumption that the drift and stochasticity can be separated, by including random noise as the input to the state equation.

Overall, this paper demonstrated that genetic programming is a viable option for symbolic discovery of SDEs, and opens the path to applying to practical applications that require robustness under high dimensionality or sparse data. Our contributions aid in understanding dynamics that exhibit noisy data, both in terms of improving the fit, as well as offering interpretability in the deterministic and stochastic components of unknown systems. In future work, integrating the proposed improvements will establish the framework as a complete and general methodology in the realm of automated scientific discovery.

\section{Acknowledgements}
This publication is part of the project ROBUST: Trustworthy AI-based Systems for Sustainable Growth with project number KICH3.L TP.20.006, which is (partly) financed by the Dutch Research Council (NWO), ASMPT, and the Dutch Ministry of Economic Affairs and Climate Policy (EZK) under the program LTP KIC 2020-2023. All content represents the opinion of the authors, which is not necessarily shared or endorsed by their respective employers and/or sponsors.

\printbibliography
\newpage

\begin{appendices}
\section{Environment details}
\label{sec:App_exp}
\paragraph{Double well problem}
The stochastic double well governs the position of a one-dimensional particle as
\begin{equation}
    \dd x = (x - x^3)\dd t + g(x) \dd w.
\end{equation}
In the deterministic case, the double well has two attracting fixed points at -1 and 1, but with additional noise the particle will move between the fixed points. Initial conditions of $x$ are sampled from $\mathcal{N}(0,0.1)$ and the trajectory is simulated until $T=50$. The double well is identified with three different diffusion terms: additive noise $g(x)=0.5$, linear multiplicative noise $g(x)=0.5 x$ and non-linear multiplicative noise $g(x)=0.5(1+x^2)$. The dataset consists of eight trajectories.

\paragraph{Van der Pol oscillator}
The van der Pol oscillator is a non-linear two-dimensional system that oscillates around zero. The stochastic dynamics are defined as
\begin{equation}
    \dd\mathbf{x}(t) = \dd \begin{bmatrix}v\\w\end{bmatrix} = \begin{bmatrix}
        w\\\mu w(1-v^2) - v
    \end{bmatrix} \dd t + \sigma\begin{bmatrix}
        0&0\\0&1+\frac{1}{2}v^2
    \end{bmatrix} \dd W
\end{equation}
with $\mu=1.0$ and $\sigma=0.2$. The initial conditions are sampled from $\mathcal{N}(0,1)$ for both dimensions. The time series are integrated until $T=50$, and eight trajectories were gathered.

\paragraph{Rössler attractor}
The Rössler attractor is a three-dimensional non-linear system that exhibits chaotic behaviour. Its stochastic dynamics are given by
\begin{equation}
    \dd\mathbf{x}(t) = \dd \begin{bmatrix}x\\y\\z\end{bmatrix} = \begin{bmatrix}
        -(y-z)\\x+ay\\b+z(x+c)
    \end{bmatrix} \dd t + \sigma\begin{bmatrix}
        x&0&0\\0&y&0\\0&0&z
    \end{bmatrix} \dd W,
\end{equation}
where $a=b=0.2$, $c=5.7$ and $\sigma=0.1$. $\mathbf{x}(0)$ is sampled from $\mathcal{N}(0, I)$, $T=50$ and the batch size is set to eight. 

\paragraph{Lorenz96 model}
The Lorenz96 model is an $N$-dimensional system, systematically changing its equations based on the number of variables. The general update of a variable $x_i$ is extended with a diffusion term, resulting in
\begin{equation}
    \dd x_i = ((x_{i+1} - x_{i-2})x_{i-1} - x_i + F)\dd t + \sigma x_i \dd W_i
\end{equation}
with $F=4$, $\sigma=0.2$ and $x_0=x_N$. The model is simulated with both five and ten variables, where the initial value of each variable is sampled from $\mathcal{N}(0,0.1)$, and the model is integrated until $T=25$ for eight trajectories. As the equation of every variable has the same general structure, the methods are only applied to learn the equations of $x_1$ in the experiments. 

\paragraph{Lotka-Volterra model}
The Lotka-Volterra model describes the interactions between prey and predator populations. The dynamics are defined as
\begin{equation}
    \dd\mathbf{x}(t) = \dd \begin{bmatrix}x\\y\end{bmatrix} = \begin{bmatrix}
        \alpha x - \beta x y\\ \delta x y - \gamma y 
    \end{bmatrix} \dd t + \sigma\begin{bmatrix}
        x&0\\0&y
    \end{bmatrix} \dd W.
\end{equation}
In our experiments, we use $\alpha=1.1$, $\beta=0.4$, $\delta=0.1$, $\gamma=0.4$, $\sigma=0.2$ and $T=50$. The initial conditions are sampled from $\mathcal{U}(5,10)$ and the dataset consists of eight trajectories.

\paragraph{Fisher-KPP equation}
The Fisher-KPP equation is a one-dimensional partial differential equation that is used to model population growth. The stochastic dynamics with multiplicative noise are defined as
\begin{equation}
    \dd u(t,x) = \left( D u_{xx} + u (1-u) \right) \dd t + \sigma u \dd W,
\end{equation}
where we use $D=1$, $\sigma=0.1$ and adopt $u_{xx}=\frac{\partial^2 u}{\partial x^2}$. Furthermore, $x\in \left[0,20\right]$ represents the coordinate space, $t\in \left[0,1\right]$ is the time span and $u(0,x)=1+\sin(x)$ is the initial condition. The boundary conditions are defined as $u(t,0)=u(t,20)$. Using the finite-differences method, the coordinate space is discretized into 64 points, with a time sampling rate of 0.02. Four trajectories were simulated in total.

\paragraph{Heat transfer}
We consider the two-dimensional heat transfer with multiplicative noise and without reaction. The noise component is scaled by the gradients with respect to the coordinate space. Hence, the dynamics follow
\begin{equation}
    \dd u(t,x,y) = D (u_{xx} + u_{yy}) \dd t + (\alpha u_x + \beta u_y) \dd W.
\end{equation}
Here, $D=0.1$, $\alpha=1$, and $\beta=1$. The coordinate space is defined by $x\in \left[0,4\right]$, $y\in \left[0,4\right]$, where both dimensions are discretized into 16 points. Furthermore, $t\in \left[0,1\right]$ with a sampling rate of 0.02, and the initial distribution of is created given $u(0,x,y)=0.2\sin(\frac{\pi}{2}x)\cos(\frac{\pi}{2}y)$. The boundary conditions are $u(t,0,y) = u(t,4,y)$ and $u(t,x,0) = u(t,x,4)$. Again, four trajectories were simulated for the dataset.

\section{Experimental details}
\label{sec:App_hyp}
The hyperparameters used by the methods in every experiments are presented in Table~\ref{Table: Hyper_params_KMSR} and Table~\ref{Table: Hyper_params_GP}.

\begin{table}[h]
\centering
\caption{Hyperparameter sets of the Kramers-Moyal expansion and sparse regression algorithm. In every run, hyperparameter optimization is performed to maximize the accuracy of the recovery. The definitions of the hyperparameters are given in Sec.~\ref{sec:baselines}.}\label{Table: Hyper_params_KMSR}\vspace{0.2cm}
\begin{small}\centering
\begin{tabular}{lccccc}
 \toprule
 \textbf{Environment} & $b$ & $\beta$ & $\alpha$ & $\lambda$ & $d$\\
 \midrule
 Double well problem    &[5, 10, 25, 50, 100]&[1, 10, 20]&[0.001, 0.01, 0.1]&[0.05, 0.1, 0.2]&3\\
 Van der Pol oscillator &[5, 10, 25, 50, 100]&[1, 10, 20]&[0.001, 0.01, 0.1]&[0.05, 0.1, 0.2]&3\\
 Rössler attractor      &[5, 10, 25]&[1, 10, 20]&[0.001, 0.01, 0.1]&[0.05, 0.1, 0.2]&2\\
 5-D Lorenz96 model     &[5, 10, 15]&[1, 10, 20]&[0.001, 0.01, 0.1]&[0.05, 0.1, 0.2]&2\\
 10-D Lorenz96 model    &[4]&[1, 10, 20]&[0.001, 0.01, 0.1]&[0.05, 0.1, 0.2]&2\\
 20-D Lorenz96 model    &[2]&[1, 10, 20]&[0.001, 0.01, 0.1]&[0.05, 0.1, 0.2]&2\\
 Lotka-Volterra model   &[5, 10, 25, 50, 100]&[1, 10, 20]&[0.001, 0.01, 0.1]&[0.05, 0.1, 0.2]&3\\
 \bottomrule
\end{tabular}
\end{small}
\end{table}

\begin{table}[h]
\centering
\caption{Hyperparameters of the genetic programming algorithm for both identifying ordinary and stochastic differential equations. The hyperparameters are defined in Sec.~\ref{sec:GP}.}\label{Table: Hyper_params_GP}\vspace{0.2cm}
\begin{small}\centering
\begin{tabular}{lcccccccc}
 \toprule
 \textbf{Environment} & $p$ & $g$ & $m$ & $s$ & $p^*$ & $g^*$ & $\eta$\\
 
 \midrule
 Double well problem    &500&50&5&15&100&15&0.1\\
 Van der Pol oscillator &500&50&5&15&100&15&0.1\\
 Rössler attractor      &500&50&5&15&100&15&0.1\\
 5-D Lorenz96 model     &1000&100&5&20&200&15&0.1\\
 10-D Lorenz96 model    &2000&200&5&20&200&15&0.1\\
 20-D Lorenz96 model    &2000&200&5&20&200&15&0.1\\
 Lotka-Volterra model   &500&50&5&15&100&15&0.1\\
 SPDEs                   &1000&50&5&15&200&15&0.1\\
 \midrule \multicolumn{8}{c}{\textbf{Multi-step genetic programming}}\\
 \midrule
 Lotka-Volterra model ($\tau=0.2$)   &2000&100&5&15&500&50&0.1\\
 Lotka-Volterra model ($\tau=0.5$)  &3000&100&5&15&500&50&0.1\\
 \bottomrule
\end{tabular}
\end{small}
\end{table}

\end{appendices}

\end{document}